%% file: main.tex
\documentclass[a4paper,conference]{IEEEtran}
\usepackage{blindtext}
\usepackage{graphicx}
\usepackage{caption}
\usepackage{hyperref}
\usepackage[dvipsnames, table, xcdraw]{xcolor}
\usepackage{standalone}
\usepackage{fancyhdr}

\begin{document}

\title{\huge{Visual Recognition of Great Ape Behaviours in the Wild}}

\author{\IEEEauthorblockN{Faizaan Sakib}
\IEEEauthorblockA{Department of Computer Science\\
University of Bristol\\
Bristol, UK\\
Email: \texttt{ss16161@bristol.ac.uk}}
\and
\IEEEauthorblockN{Dr. Tilo Burghardt}
\IEEEauthorblockA{Department of Computer Science\\
University of Bristol\\
Bristol, UK\\
Email: \texttt{tilo@cs.bris.ac.uk}}}

\maketitle

\input{abstract}

\IEEEpeerreviewmaketitle

\section{Introduction} % 1 page

\input{introduction}

\section{Dataset and Annotations} % 1 page

\input{dataset}

% Both sections below should be 2 pages

\section{Methods} % 1 page
\label{sec:method}

\input{methods}

\section{Experiments and Basic Results} % 1 page

\input{experiments}

\section{Cross-validated Results} % 1 page

\input{results}

% Includes all-encompassing result table

\section{Conclusion} % 1 page

\input{conclusion}


\bibliography{bibliography}
\bibliographystyle{IEEEtran}

\end{document}

%% file: abstract.tex
\begin{abstract}
    Whilst species detection~\cite{apedetection} and individual identification~\cite{chimprecognition} of great apes have previously been attempted with success, research into automated behaviour recognition remains a human-centric task~\cite{survey}. In this paper we present a first great ape-specific visual behaviour recognition system and related data annotations for detecting core ape behaviours. Among others, these include sitting, walking, and climbing. The presented basic dual-stream model with late fusion is capable of performing multi-subject multi-behaviour recognition on apes in challenging camera trap footage. More than 180,000 frames across 500 videos from the PanAfrican dataset~\cite{panafrican} were annotated with individual IDs and behaviour labels to end-to-end train and evaluate the system. In summary, our key contributions include a proposed system capable of an accuracy of 73.52\%, along with the behaviour annotated dataset, key code and network weights.
\end{abstract}

%% file: introduction.tex
The task of behaviour recognition is concerned with the automatic identification of a subject's activity within a video. Deep convolutional neural networks (CNNs) have proven to be proficient in performing this task. However, research in this area has remained largely focused on human activity~\cite{survey}. Similarly, current behaviour-annotated datasets do not extend beyond humans as its subjects~\cite{kinetics, ucf101}.

% Animal biometric system
Yet, deep learning clearly has the potential to play a key role in substantially enhancing efforts of research and conservation of great apes. The foundations of great ape ``biometrics''~\cite{animalbiometrics} are already in place, with effective methods for ape detection~\cite{apedetection} and identification~\cite{chimprecognition} having been established. The addition of behaviour recognition represents the next step towards a system that can truly automate the monitoring of great apes. 

%% file: dataset.tex
\begin{figure}[t]
  \includegraphics[width=0.48\textwidth]{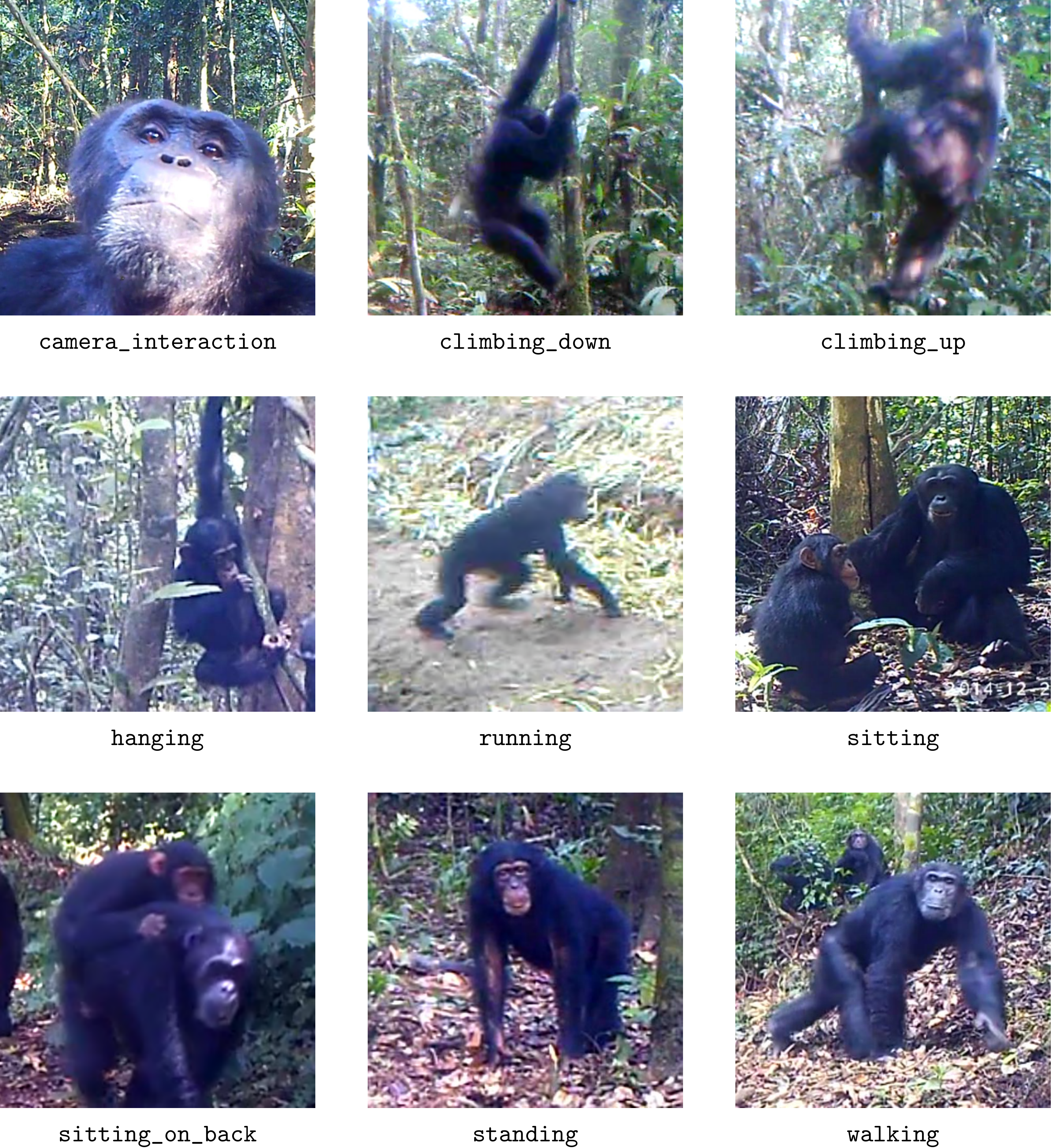}
  \caption{\textbf{Great Ape Behaviours.} Examples of the nine core behaviours targeted in this paper. We produce a large, manually annotated training data corpus for networks to learn these behaviours and evaluate a basic dual-stream CNN architecture to demonstrate the viability of automated behaviour detection.}
\label{fig:title}
\end{figure}

\begin{figure*}[t]
  \includegraphics[width=\textwidth]{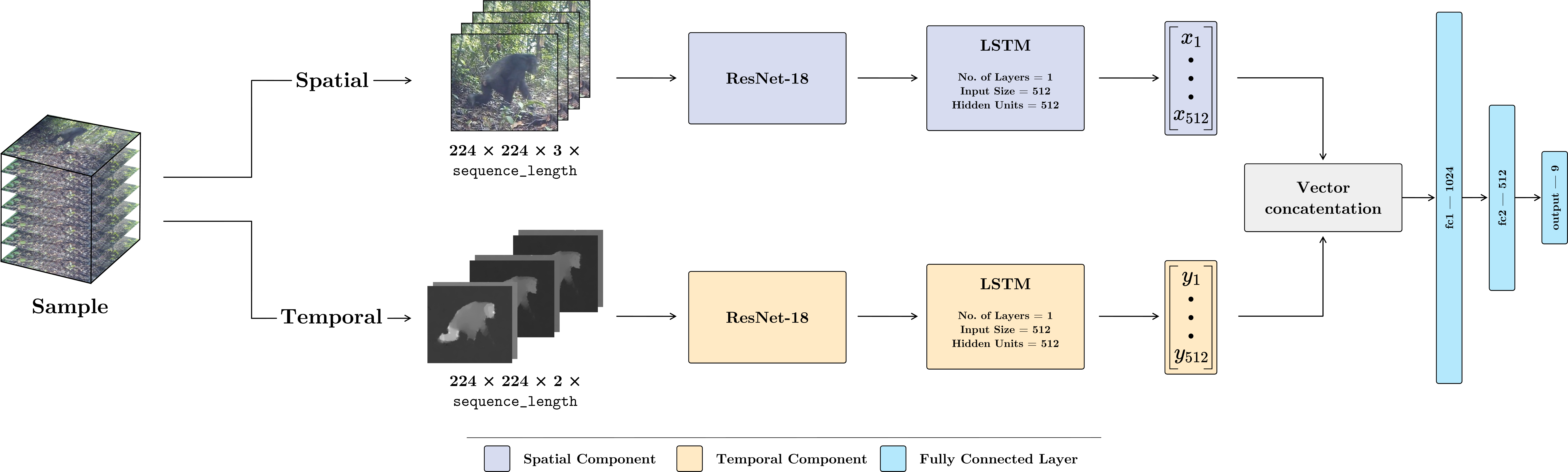}
  \caption{\textbf{Two-Stream ResNet-18 + LSTM great ape behaviour recognition model architecture:} The spatial stream takes in a sequence of RGB frames and the temporal stream takes in a sequence of greyscale optical flow frames. An LSTM network is appended to each of the ResNet-18 CNNs. The vector output of size 512 from both LSTM networks are fused using concatenation. This is followed by a two layer fully connected network to produce the final classification scores.}
  \label{fig:arch}
\end{figure*}

% Which dataset was used
We utilise the PanAfrican~\cite{panafrican} dataset as an example of 'in the wild' capture of great ape behaviour. It is one of the largest of its kind in existence, made up of thousands of hours of jungle trap camera video footage filmed in central Africa. There are two species of great apes that feature in the subset of data we use: gorillas and chimpanzees. At any given frame, the total number of apes simultaneously present may range from 0 to 8. The apes across the dataset vary in species, age, gender and size. The videos span across multiple locations, at varying time of day.

% Partly annotated for detection research
In this work we extend the dataset produced in~\cite{apedetection} which includes ape objects with manually annotated bounding boxes. The augmented annotations created as part of this work can be found at \url{data.bris.ac.uk} under the name PanAfrican2020. In total there are $\sim180,000$ \texttt{XML} annotation files, individually mapping to every frame spanning across 500 videos captured at 24~frames per seconds. We split the videos into training, validation and test sets at a ratio of 400:25:75. Scripts were written to automate the process of labelling where possible. We extended each ape object with the following metadata:

\begin{itemize}
    \item \textbf{Behaviour:} The activity (e.g. walking) being exhibited by the great ape. This is made up of one behaviour, as simultaneous behaviours are not taken into account.
    \item \textbf{Identification number (ID):} A unique integer to identify each ape within a video, assigned by order of appearance starting from 0.
\end{itemize}

A total of nine distinguishable core behaviours were identified and annotated: \texttt{camera\_interaction}, \texttt{climbing\_down}, \texttt{climbing\_up}, \texttt{hanging}, \texttt{running}, \texttt{sitting}, \texttt{sitting\_on\_back} (a child ape mounted on an older ape's back), \texttt{standing} and \texttt{walking} as shown in Fig.~\ref{fig:title}. Meanwhile, the ID label establishes tracklets for each ape present throughout the duration of a video.

Note that there were “auxiliary” behaviours, namely eating and the act of scavenging, that were usually seen while other behaviours were already being exhibited (e.g. an ape could eat while sitting or walking). However, these were not considered as core behaviour classes since the model defined for this project aimed to only classify one behaviour at any one time.

%% file: methods.tex
% Two-stream
A two-stream approach, akin to the work of Simonyan et al.~\cite{twostream}, was utilised as the backbone of the behaviour recognition model. It consists of two CNNs taking into account the spatial and temporal dependencies of a video separately~(see~Fig.~\ref{fig:arch}). They take in a sequence of $n$ consecutive RGB frames and $n$ grayscale optical flow images respectively.

% Data sampling
To fulfil the multi-subject multi-behaviour requirements of the model, we consider sampling every ape present in a given video. A \textit{sequence length} of 20 consecutive frames is used as input to both streams. A \textit{sampling rate} of 20 is used, effectively leaving no interval between sequences sampled. Finally, a \textit{behaviour duration threshold} is considered, signifying the number of consecutive frames an ape has to exhibit a behaviour for it to qualify for sampling. The threshold helps filter the data to obtain quality samples showing sustained displays of behaviour where prominent characteristics are likely to be exhibited. We use a threshold of 72 frames, equalling 3 seconds.

For each valid sample, we crop the bounding boxes around the ape as per the specified ID before assembling them into sequences to be used as input to the model.

Features are then extracted from the sequence input with the use of ResNet-18~\cite{resnet} CNNs. Both spatial and temporal ResNet-18 CNNs are pretrained on ImageNet~\cite{imagenet} and subsequently fine-tuned on our dataset to reap the benefits of transfer learning. Long short-term memory (LSTM) networks are appended to the CNN of each stream. The feature map of size 512 output from each of the CNNs are used as input for the LSTM networks.

% Fusing before using 1 LSTM had poor performance possible due to the inconsistency created in the timesteps where one timestep was a spatial frame and the following was temporal and so on.

% Use of detectors in real-world
In the case where there is no tracklet information present, detectors such as YOLO~\cite{yolov3} can be easily trained and integrated. Additionally, the domain-specific system for ape detection proposed by Xinyu et al.~\cite{apedetection} is a viable option.

% Vector concatenation and fully connected layer
The output from each of the LSTM networks are fused with vector concatenation amounting to a late fusion approach. Finally, this combined vector is passed through a two-layer fully connected network to obtain classification scores.

% Balanced sampling
Balanced sampling was used to address class imbalance found in the dataset, by ensuring every training batch consists an equal number of samples for each class. Additionally, focal loss~\cite{focal} is used as an alternative to cross-entropy loss.

% Hyperparameters
The learning rate is set to 0.0001. $L_2$ regularisation of 0.01 is used. The focal loss is used with parameters $\alpha$ = 1.00 and $\gamma$ = 1.00. The batch size is set to 9 (reduced to the lowest size possible while using balanced sampling to counter the high memory requirements of the LSTM networks). Both LSTM networks use 1 layer and 512 hidden units each. The SGD optimiser is used with momentum set to 0.9.

%% file: experiments.tex
\begin{figure}[t]
  \includegraphics[width=0.45\textwidth]{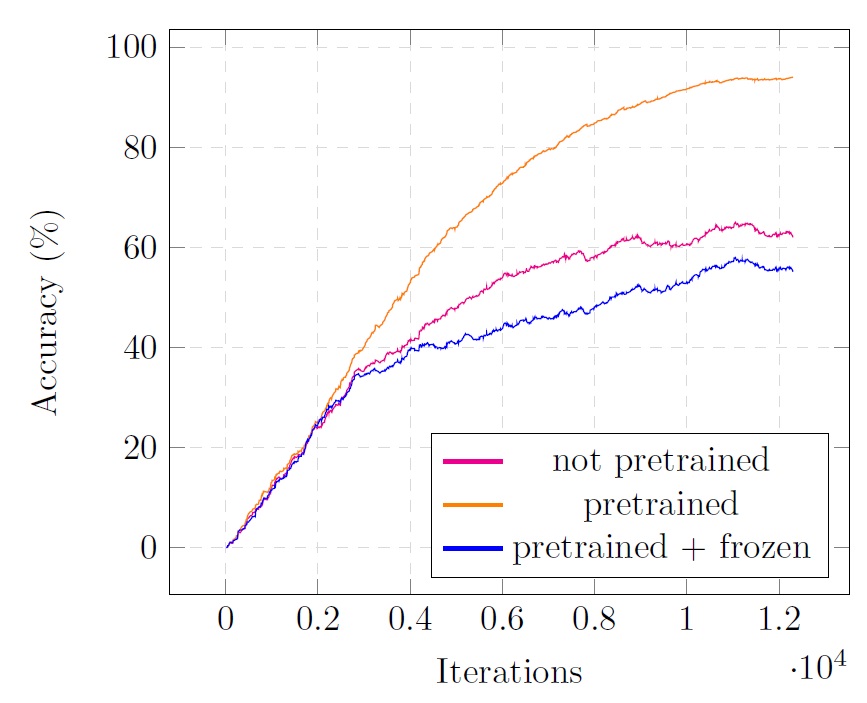}
  \caption{\textbf{Transferability of ImageNet Weights to Spatial Stream of Recognition Model.} The graph shows accuracy during a training run with and without using pretrained weights from ImageNet to initialise the spatial stream of the network architecture. The large gains in training accuracy with the use of pretrained weights can be observed.}
  \label{fig:final}
\end{figure}

\textbf{Simple Baseline Model.} Initially, a baseline model was tested with the use of ResNet-18 CNNs pretrained on ImageNet~\cite{imagenet} in a two-stream architecture. We found that ResNet-18 proved to show good performance while retaining a relatively low level of complexity, using only \textbf{$\sim$8.7\%} of parameters when compared to VGG-16. In addition, the pretrained weights were clearly beneficial for the training process due to an increased final accuracy. The impact of pretraining on the spatial stream appeared to be much greater than on the temporal stream. Following Feichtenhofer et al.~\cite{twostreamfusion} we tested both late fusion and convolutional fusion to achieve dual-stream convergence, where convolutional fusion comprehensively captured spatio-temporal dependencies leading to overall better performance. This baseline model featuring the design choices described above performed with a Top1 accuracy of \textbf{61.39\%}, Top3 accuracy of \textbf{86.34\%}. 

\textbf{Additional Features.} Balanced sampling increased this Top1 accuracy by \textbf{7.39\%}, and Top3 accuracy by \textbf{6.83\%}. The parameters of the focal loss were tested, with best performance found where $\alpha=1.00$ and $\gamma=1.00$. LSTM networks were introduced to the model as an alternative method in capturing multi-frame information. The use of 2 LSTMs using 1 layer each without dropout performed best placing the units after the ResNet-18 CNN of each stream~(see~Fig.~\ref{fig:arch}). Since the LSTMs output 1D vectors, using convolutional fusion which involves 3D convolutional layers was no longer suitable. Thus, we reverted back to using late fusion upon the concatenated LSTM output vectors. Table \ref{tab:validationresults} shows the evaluation of the two models against the withheld test set, indicating the significant improvement in performance on the baseline model.

An attempt to use only one LSTM in the model for lower memory consumption was  made by performing pre-LSTM fusion between the two streams. However, this resulted in a negative impact on performance. An alternate approach could be to perform spatial and temporal feature extraction using an I3D model~\cite{i3d} before training a single LSTM with its output.

\begin{table}[h]
\centering
\resizebox{0.4\textwidth}{!}{%
\begin{tabular}{|c|c|c|}
\hline
\multicolumn{3}{|c|}{\cellcolor[HTML]{EFEFEF}\textbf{Single Train-Test Split Results}} \\ \hline
Model                & Top1 Accuracy          & Top3 Accuracy              \\ \hline
Baseline             & 61.39\%        & 86.34\%                 \\ \hline
Optimised            & \textbf{74.82}\%        & \textbf{95.49}\%                   \\ \hline
\end{tabular}%
}
\caption{\textbf{Single Train-Test Split Results.} The introduction of balanced sampling, focal loss and LSTM into the baseline architecture leads to a significant performance increase.}
\label{tab:validationresults}
\vspace{10pt}
\end{table}

We experimented with a sequence length of $5$, $10$, $20$, $30$ frames. We were limited to 30 frames due to memory constraints. The model's performance with a sequence length of 20 frames was found to be the best, as a subsequent increase to 30 frames showed a significant drop in performance.

The optimised model with additional features performed with a Top1 accuracy of \textbf{74.82\%}, Top3 accuracy of \textbf{95.49\%}.
 
We also report preliminary detection results with YOLOv3~\cite{yolov3}, performing with a mAP of \textbf{81.27\%} on the test set. Results of the ape detector proposed by Xinyu et al. on the same dataset can be found in the original paper~\cite{apedetection}.

%% file: results.tex
We evaluated the model described in Section \ref{sec:method} using 4-fold cross validation. Each fold consisted of 375 training videos and 125 test videos. Table \ref{tab:finalresults} displays the resulting performance measures. It can be seen that the performance seen on a single train-test split can be widely retained, suggesting the model generalises reasonably within the domain tested.

\begin{table}[h]
    \centering
    \resizebox{0.35\textwidth}{!}{%
    \begin{tabular}{|
    >{\columncolor[HTML]{FFFFFF}}c |
    >{\columncolor[HTML]{FFFFFF}}c |}
    \hline
    \multicolumn{2}{|c|}{\cellcolor[HTML]{EFEFEF}\textbf{Final Cross Validated Results}}                             \\ \hline
    Top1 Accuracy                             & Top3 Accuracy                     \\ \hline
    \textbf{73.52\%} & \textbf{94.07\%} \\ \hline
    \end{tabular}%
    }
    \caption{\textbf{Final 4-fold Cross-validated Results.} The performance of the single train-test split can be widely retained indicating good generalisability of the model to unseen data.}
    \label{tab:finalresults}
\end{table}

\section{Qualitative Analysis}

\begin{figure}[t]
  \includegraphics[width=0.48\textwidth]{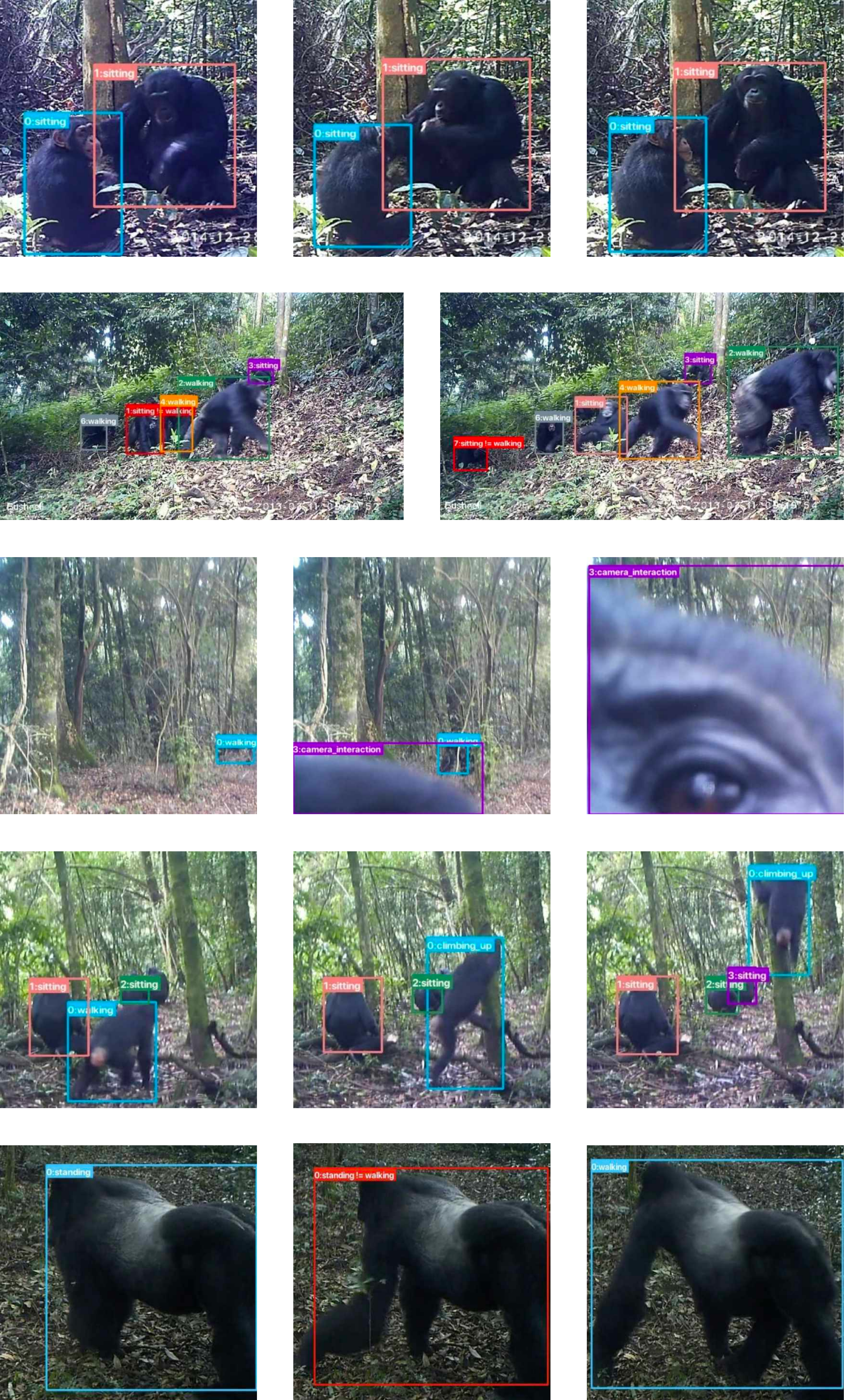}
  \caption{\textbf{Examples Output of the Model.} Rows show video skim sequences from the test set. Each ape is indicated by its bounding box, with its ID and detected behaviour labelled on the top left. A red box signifies an incorrect prediction, with the true behaviour appended on the label. The bottom row shows an instance of a transitional behaviour where the model fails to recognise the start of the ape walking in time.}
  \label{fig:exampleoutput}
\end{figure}

When visually observing the model's predictions, we identified a number of trends where the model performs poorly. One of these was identified in the form of transitional behaviours. They are ambiguous periods where an ape is in the process of ending its current behaviour and beginning to display the next. Fig.~\ref{fig:exampleoutput}~(bottom row) illustrates an example skim of such a transition.

The model is often unable to predict the behaviour of \texttt{running}, primarily mistaking it as \texttt{walking}. This suggests that the model may be ineffective in picking up the difference in speed between the two behaviours. There are also occasions where an ape stands stationary, while using their arms to dig or pick objects up. This movement, while standing, closely resembles the initial motions of walking. As such, the model predicts that the ape is starting to walk, when it is not.

%% file: conclusion.tex
In this paper we introduced the first bounding-box level behaviour meta-dataset for great apes covering 180,000 frames across 500 videos of the PanAfrican camera trap archive~\cite{panafrican}. We evaluated a basic two-stream deep learning model as a baseline using this challenging footage taken in the wild. By exploring the use of balanced sampling, focal loss and LSTMs we were able to significantly improve results. The outcome of a 4-fold cross validation procedure shows that the final model is capable of classifying nine core behaviours among gorillas and chimpanzees, at a Top1 accuracy of \textbf{73.52\%}. Future work will focus on cross-fertilisation of behaviour detection with recent advances in great ape pose recognition~\cite{2020_CVPR}.

We hope that the behaviour annotations and proof of concept provided in this paper can aid research of the biological and conservational sciences in order to further our understanding and ultimately the protection of these charismatic species. 

\section*{Acknowledgements}
We would like to thank the entire team of the Pan African Programme: ‘The Cultured Chimpanzee’ and its collaborators for allowing the use of their data for this paper. Please contact the copyright holder Pan African Programme at \url{http://panafrican.eva.mpg.de} to obtain the videos used from the dataset. Particularly, we thank: H Kuehl, C Boesch, M Arandjelovic, and P Dieguez. We would also like to thank: K Zuberbuehler, K Corogenes, E Normand, V Vergnes, A Meier, J Lapuente, D Dowd, S Jones, V Leinert, EWessling, H Eshuis, K Langergraber, S Angedakin, S Marrocoli, K Dierks, T C Hicks, J Hart, K Lee, and M Murai. Thanks also to the team at https://www.chimpandsee.org. The work that allowed for the collection of the dataset was funded by the Max Planck Society, Max Planck Society Innovation Fund, and Heinz L. Krekeler. In this respect we would also like to thank: Foundation Ministre de la Recherche Scientifique, and Ministre des Eaux et Forłts in Cote d’Ivoire; Institut Congolais pour la Conservation de la Nature and Ministre de la Recherch Scientifique in DR Congo; Forestry Development Authority in Liberia; Direction des Eaux, Forłts Chasses et de la Conservation des Sols, Senegal; and Uganda National Council for Science and Technology, Uganda Wildlife Authority, National Forestry Authority in Uganda. 